\documentclass[11pt]{article}
\usepackage[]{acl}
\usepackage{times}
\usepackage{latexsym}
\usepackage{booktabs}
\usepackage{graphicx}
\usepackage[T1]{fontenc}
\usepackage[utf8]{inputenc}
\usepackage{microtype}
\usepackage{inconsolata}
\usepackage{xurl}

\usepackage{eso-pic}
\usepackage{xcolor}
\usepackage{amsmath}

\title{Counting Documents Is Not Counting Text:\\Unit Bias in Web-PDF Corpus Statistics}

\author{Luca Foppiano \\
  Common Crawl Foundation \\
  \texttt{luca@commoncrawl.org}
}

\begin{document}
\maketitle

\begin{abstract}
PDF corpora advertise their size in tokens, but every rate they publish (coverage, OCR routing, re-fetch recovery, language mix) is computed per document, and none decomposes its token total.
Because PDF length is extremely skewed, the two units can describe the same corpus very differently.
We ask how the headline statistics of a web-PDF corpus change when each document is weighted by the text it contributes rather than counted once.
We used \texttt{CC-MAIN-2021-31-PDF-UNTRUNCATED} (7.9M Common Crawl PDFs, 32.6B tokens), the one public corpus that pairs the fragments Common Crawl stored with the re-fetched originals.
Text mass is highly concentrated: 3.02\% of text-bearing documents hold half the tokens (Gini 0.807).
The clearest consequence is Common Crawl's payload cap, which truncated 23.06\% of these documents but 63.08\% of their text.
Reconstructing the truncated fragments and extracting both versions, two widely used text-layer parsers recover only 1.4\% and 11.4\% of that exposed text, so roughly 55--62\% of the corpus's text is unrecoverable from the crawl by such pipelines; under the 5\,MiB cap adopted in March 2025, 30.19\% of tokens would still be exposed.
We recommend that corpus statistics be reported in both units, documents and tokens.
\end{abstract}

\section{Introduction}

PDFs have become a first-class source of pretraining text \citep{olmocr2025,pdfa2024}.
In one current open data pool, PDF-derived text amounts to roughly 4.3T tokens \citep{marin2025datakit}.\footnote{Marin's \texttt{datakit} source registry: FinePDFs 1{,}186.5B English plus 1{,}353.4B across 18 non-English subsets, and \texttt{dolma4pdfs} 1{,}804.0B over 137{,}132{,}279 documents. \url{https://github.com/marin-community/marin}} FinePDFs alone is described as ``about 3 trillion tokens across 475 million documents'' \citep{finepdfs2025}.
How do a web-PDF corpus's headline statistics change when each document is weighted by the text it contributes rather than counted once?

Those corpora are not naive about tokens: they headline them.
What none of them does is compute a \emph{rate} in tokens, or break a token total down.
FinePDFs routes 368.8M of 1.29B \emph{files} to OCR and recovers 53.5\% of its truncated \emph{files} by re-fetching \citep{finepdfs2025}; CCpdf reports per-\emph{document} success rates \citep{turski2023ccpdf}; PDFA gives three totals (2{,}159{,}432 documents, 18M pages, 9.7B tokens) but no drop rates in any unit \citep{pdfa2024}.
No PDF corpus we are aware of decomposes its token total by length or quality.

This would be harmless if documents were interchangeable. They are not.
The web-text community already recognises this: Nemotron-CC reports a \emph{token}-weighted yield, ``+57.4\% more high-quality tokens'' \citep{su2025nemotron}, and PDFs have far more extreme length skew than HTML.
FinePDFs itself observes that its documents have a ``$\sim$5.3k character median (about 2$\times$ other corpora)'' with a ``95th percentile $\sim$68k'' against ``$\sim$11--13k elsewhere'' \citep{finepdfs2025}, and then reports every rate per document anyway.

We measure what those rates become when weighted by text.
The shipped metadata of \texttt{CC-MAIN-2021-31-PDF-UNTRUNCATED} lets us re-weight three of the quantities above, document length, producer toolchain and truncation; the argument applies equally to OCR routing, filter drop and language mix, which we do not measure here.
Our contributions are:

\begin{enumerate}
  \setlength{\itemsep}{0pt}
  \item the first token-weighted characterisation of a web-PDF corpus, with a direct document-versus-token comparison of every headline statistic (\S\ref{sec:conc}--\ref{sec:tool});
  \item a concentration result: text mass in web PDFs is far more skewed than document counts suggest (Gini 0.807);
  \item the first measurement of what Common Crawl's truncation cap removes in text: 23.06\% of documents but 63.08\% of tokens, of which two text-layer parsers recover 1.4--11.4\%, with 30.19\% still exposed at the new 5\,MiB limit (\S\ref{sec:trunc}--\ref{sec:cap5}).
\end{enumerate}

\section{Related Work}
\label{sec:related}

\paragraph{Totals in tokens, rates in documents.}
FinePDFs \citep{finepdfs2025} routes documents between a text path (Docling) and a GPU OCR path with a learned classifier reporting F1 0.71 on the OCR class, and quantifies every stage in files.
CCpdf \citep{turski2023ccpdf} tabulates ``number of documents per processing step and language'' and reports no token count for its own corpus.
PDFA \citep{pdfa2024}, derived from the same corpus we study, filters per document, discarding files over 100\,MB or slower than 500\,ms to render, i.e.\ selecting on \emph{size}, and publishes no drop rates.
GovScape~\citep{huang2026govscapepublicmultimodalsearch} excludes ``pages with no text embedded within the PDF'' without quantifying the exclusion in any unit.
Extraction benchmarks \citep{omnidocbench2024,olmocr2025} measure fidelity on documents that parsed and report no coverage at all.
\citet{kuznetsov2023inclusive} argue more generally that what counts as ``text'' is under-specified in NLP.
In language-model research the question has been settled in practice rather than in principle: the token is the unit in which corpora are sized, data mixes are set and compute is scaled, and it is the unit a model consumes.
Corpus statistics have not followed consistently.

\paragraph{Truncation is named but never quantified.}
Common Crawl caps the payload it stores per record.
PDFs are the format worst affected: for the crawl \texttt{CC-MAIN-2023-06} they were 0.8\% of successfully fetched records but 11.85\% of 88\,TiB of WARC storage \citep{pdfa2023blogpost}.
CCpdf stated the consequence plainly in 2023: the crawler's 1\,MB cap is ``quite high for HTML pages, but unfortunately rather low for PDF files'' \citep{turski2023ccpdf}.
Their workaround was to re-download the truncated documents from origin; to our knowledge nobody since has quantified the effect.



From \texttt{CC-MAIN-2025-13} (released in March 2025) the cap rose from 1\,MiB to 5\,MiB, reported as +13\% fetched content (403\,TiB $\rightarrow$ 455\,TiB) \citep{cc2025march}.

The Common Crawl Foundation reported the effect of the change on truncation by MIME type: PDFs fell from 25.7\% to 6.8\%, HTML from 2.2\% to 0.04\%. 
On average, all types fell from 2.25\% to 0.14\%.
Even after the change PDFs are truncated at roughly 49$\times$ the rate of content generally.
What is not reported, before or after, is how much text that removes.

\section{Data and Method}
\label{sec:data}

\paragraph{Corpus.}
\texttt{CC-MAIN-2021-31-PDF-UNTRUNCATED} \citep{safedocs2021} contains every PDF found in Common Crawl \texttt{CC-MAIN-2021-31}, with payloads that Common Crawl truncated \emph{refetched whole} from origin.
It ships five metadata tables covering 8.3M URLs.
Because \texttt{file\_name} is the post-SHA-256 identity, we deduplicate to one row per file, giving 7{,}932{,}654 unique documents.

\paragraph{The unit.}
Text mass is taken from \texttt{tika\_eval\_num\_tokens}, a token count already published as part of the corpus metadata \citep{safedocs2021} (Apache Tika 2.8.0, tesseract disabled), a whitespace/ICU count rather than an LLM tokenizer's.
The total is 32{,}570{,}135{,}761 tokens.
We claim only proportions, which is all the argument requires.
Tika stops counting at 10{,}000{,}000 tokens, so a longer document is recorded at that value; only two documents in the corpus reach it, and both are therefore undercounted.

\paragraph{Measures.}
For any subset $S$ of the $N$ documents we report its \emph{document share} $|S|/N$ and its \emph{token share} $\sum_{d\in S} t_d / \sum_{d} t_d$, where $t_d$ is the token count of document $d$; the ``ratio'' column of Table~\ref{tab:shape} is the second divided by the first.
The \emph{head share} at level $q$ is the smallest fraction of documents, taken in decreasing order of $t_d$, whose tokens reach a fraction $q$ of the total.
Concentration is summarised by the Gini coefficient in its standard form: with the $n$ documents sorted so that $t_{(1)}\le\dots\le t_{(n)}$,
\begin{equation}
G \;=\; \frac{2\sum_{i=1}^{n} i\,t_{(i)}}{n\sum_{i=1}^{n} t_{(i)}} \;-\; \frac{n+1}{n}.
\label{eq:gini}
\end{equation}
$G=0$ means every document carries the same amount of text and $G\to1$ means all of it sits in one document; values around 0.3--0.4 are typical of national income distributions, the setting the coefficient comes from.
We compute it over the $n=7{,}292{,}093$ documents with at least one token, so the 8.1\% of zero-token documents do not inflate it, and the two censored documents can only lower it.
Figure~\ref{fig:lorenz} shows the underlying Lorenz curve.
The categories of Table~\ref{tab:shape} are bins on page count and first-page orientation, chosen so that the two extremes, single-page files and files over 50 pages, are unambiguous; they are not genre labels, and a long form is binned with long reports.

\paragraph{Exposure, recovery, loss.}
Three quantities recur in \S\ref{sec:trunc}--\ref{sec:cap5}.
\emph{Exposure} $E$ is the token share of the documents Common Crawl truncated; it is computed from the shipped metadata over all 7.9M documents.
\emph{Recovery} $r_{e,d}$ of a truncated document $d$ under extractor $e$ is the number of tokens $e$ extracts from the truncated document, divided by the number it extracts from the whole document.
The corpus rate $r_e$ sums both quantities over the documents processed before dividing, so long documents weigh more.
It is measured by re-creating the fragments (\S\ref{sec:sim}).
\emph{Loss} is $E\,(1-r_e)$: the share of the corpus's text that a pipeline reading the text layer with $e$ cannot obtain from the crawl.
Loss is an estimate under two assumptions, both tested in \S\ref{sec:sim}: that recovery on the documents we processed transfers to the exposed mass as a whole, and that a document's weight is the same whether counted in Tika tokens or in the whitespace tokens of our own extraction.

\paragraph{Truncation.}
The provenance table records \texttt{cc\_truncated}, \texttt{fetched\_status}, \texttt{fetched\_length} and the WARC byte range per file (\texttt{cc\_truncated=`length'} coincides exactly with \texttt{fetched\_status=`REFETCHED\_SUCCESS'}).
For truncated records the WARC record length pins to $\sim$1.049\,MB, so the cap is directly observed, and the re-fetched originals give the true size of every document Common Crawl stored only as a fragment: a pairing no other public corpus offers.

\section{Results}

\subsection{Text mass is extremely concentrated}
\label{sec:conc}

\begin{table}[t]
\centering\small
\begin{tabular}{lrrr}
\toprule
Bin & doc \% & tok \% & ratio \\
\midrule
$>$50\,pp & 4.42 & \textbf{49.71} & \textbf{11.24}$\times$ \\
4--30\,pp & 31.64 & 26.87 & 0.85$\times$ \\
31--50\,pp & 3.21 & 9.08 & 2.83$\times$ \\
landscape, any length & 14.08 & 7.76 & 0.55$\times$ \\
2--3\,pp & 21.20 & 3.76 & 0.18$\times$ \\
1\,pp & 24.96 & 2.70 & \textbf{0.11}$\times$ \\
\midrule
$>$50\,pp, any orientation & 5.00 & 53.53 & 10.71$\times$ \\
\bottomrule
\end{tabular}
\caption{The same corpus counted two ways. Rows above the rule are mutually exclusive bins on page count and first-page orientation; orientation takes precedence, so a long landscape document falls in the landscape row and the page-count rows cover portrait documents only. The memo row below the rule ignores orientation and is the threshold used in the text and in \S\ref{sec:robust}.} 0.49\% of documents with unparseable page metadata are omitted.
\label{tab:shape}
\end{table}

\begin{figure}[t]
\centering
\includegraphics[width=\columnwidth]{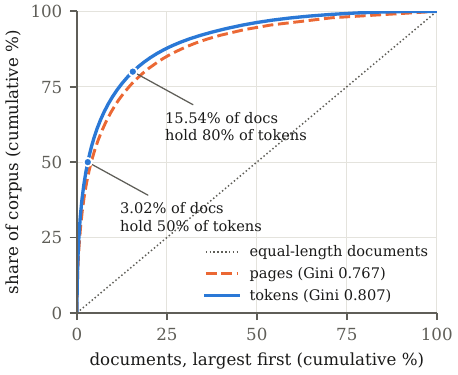}
\caption{Cumulative share of tokens (solid) and of pages (dashed) held by the largest documents, over the 7{,}292{,}093 text-bearing documents and the documents with a page count respectively. 
The marked points are the head shares quoted in the text: 3.02\% of documents hold 50\% of the tokens and 15.54\% hold 80\%. The dotted diagonal is a corpus in which every document has the same length.}
\label{fig:lorenz}
\end{figure}

Table~\ref{tab:shape} gives the page-count distribution in both units.
Counting by page count alone, regardless of orientation (memo row of Table~\ref{tab:shape}), documents over 50 pages are 5.00\% of the corpus and 53.53\% of its text: a document in that tail carries about 10.7$\times$ an average document's text. We use this threshold for the concentration claim here and in \S\ref{sec:robust}.
The mirror image, from the two bottom rows above the rule: the 46.2\% of documents with three pages or fewer contribute 6.46\% of the text.
Over the 7{,}292{,}093 text-bearing documents, 3.02\% hold half the tokens and 15.54\% hold 80\% (Figure~\ref{fig:lorenz}); the Gini coefficient of tokens across documents (Eq.~\ref{eq:gini}) is 0.807.

Counted by document, the corpus is dominated by files of three pages or fewer; counted by text, it is a corpus of long documents.
Both descriptions are arithmetically correct.

\subsection{The concentration is not an artifact of one extractor}
\label{sec:robust}

The concentration in \S\ref{sec:conc} rests on one extractor and one way of counting: Tika's text extraction and its whitespace token count. 
Either could favour long documents, the first if short files are more often scans or forms that yield little text, the second because the count depends on script and layout. 
We therefore validate the distribution against page counts, which the corpus authors extracted with Poppler, an independent tool measuring an entirely different quantity.
Page counts reproduce the same concentration: the Gini coefficient of \emph{pages} across documents is 0.767 against 0.807 for tokens, 3.98\% of documents hold half the pages against 3.02\% for tokens, and documents over 50 pages account for 54.27\% of all pages against 53.53\% of all tokens, a difference of 0.74 percentage points.
Two tools measuring two different quantities give the same answer: a shared extraction bias cannot produce this agreement.

\subsection{The \TeX{} share more than doubles}
\label{sec:tool}

Classifying the \texttt{producer}/\texttt{creator} strings into toolchain families, a \TeX{} toolchain accounts for 1.66\% of documents but 4.05\% of tokens (2.43$\times$).
PowerPoint runs the other way, 2.24\% of documents against 0.82\% of tokens (0.37$\times$), and Microsoft Word is 24.95\% of documents against 19.37\% of tokens.

Toolchain is a proxy for the production pipeline, not for genre: Word and PowerPoint also carry scholarly text, and producer strings are self-reported and incomplete (see Limitations).
What the comparison shows is that the \TeX{} share differs by 2.4$\times$ between the two units, and the token unit is the one a language model consumes.




\subsection{Truncation: 23\% of documents, 63\% of the text}
\label{sec:trunc}

This section measures exposure, the token share of the documents Common Crawl truncated; \S\ref{sec:sim} measures how much of it is recoverable.

\begin{table}[t]
\centering\small
\begin{tabular}{lrr}
\toprule
& docs \% & \textbf{tokens \%} \\
\midrule
Truncated by CC (1\,MiB) & 23.06 & \textbf{63.08} \\
\ \ still truncated at 5\,MiB & 5.94 & \textbf{30.19} \\
\bottomrule
\end{tabular}
\caption{Common Crawl's truncation cap, expressed in both units, over the 7{,}932{,}878 files with provenance records.}
\label{tab:trunc}
\end{table}

Common Crawl truncated 1{,}829{,}061 of this corpus's documents at the 1\,MiB cap in force at the time, corresponding to 23.06\% of them.
Those documents hold 63.08\% of the corpus's text (Table~\ref{tab:trunc}).
Truncated documents carry 2.74$\times$ the mean token mass; their median true size is 2.62\,MB against 190\,KB for the rest.

Applying the current 5\,MiB cap to the corpus, 5.94\% of documents and 30.19\% of tokens would still be truncated.
The March 2025 change (\S\ref{sec:related}) therefore removes roughly half of the exposure and leaves the other half in place.
Our document-level counterfactual is consistent with Common Crawl's own post-change measurement of 6.8\% on 2025 crawls  \citep{cc2025march}, computed on a population five years younger; the token figure has no published counterpart.

In bytes, the 1\,MiB cap kept 1.738\,TiB of these files' 9.449\,TiB total, discarding 81.6\%; storing them whole would have cost 7.712\,TiB of additional archive.
The 5\,MiB cap lands between the two: it would keep 5.106\,TiB (46.0\% discarded), so the March 2025 change spends 3.368\,TiB of that 7.712\,TiB and leaves 4.344\,TiB behind.
At either cap, byte loss overstates text loss: 81.6\% of bytes against 63.08\% of tokens at 1\,MiB, 46.0\% against 30.19\% at 5\,MiB, because PDF bytes are largely images and embedded fonts.
The byte figure measures the archive; the token figure measures the corpus.

\subsection{Little of the exposed text is recoverable}
\label{sec:sim}

This section measures recovery, as defined in \S\ref{sec:data}, and combines it with the exposure of \S\ref{sec:trunc} into loss.
To do so, we reconstruct what Common Crawl held, cutting each of this corpus's 1{,}829{,}061 truncated documents at 1\,MiB, and extract text from both the fragment and the whole file.
Whether a fragment is recoverable may be a property of the parser rather than of the file, so we run two independent engines, PyMuPDF (MuPDF) \citep{pymupdf} and PDFium (Chromium's) \citep{pypdfium2}, and report on the 1{,}225{,}130 documents both tools successfully processed, and whose whole version yields text.

PyMuPDF recovers 11.4\% of the tokens and 20.7\% of the pages; PDFium recovers 1.4\% and 1.9\%, a gap of 8.4$\times$ on identical input.
The two agree to within 1.7\% on the \emph{intact} versions of those same documents, so the divergence is specific to damaged input and is not a general difference in extraction quality.
There is therefore no single ``recovery rate'' for a truncated PDF: the quantity is not defined until the extractor is named.

The shape of the failure differs as sharply as its size, and both shapes are invisible to a pipeline that counts successful parses.
PyMuPDF opens 98.9\% of truncated files, because MuPDF rebuilds a missing cross-reference table, but the page content streams lie beyond the cut, so the document opens, reports a page count, and returns no text: 72.4\% yield nothing at all.
PDFium attempts no such reconstruction and refuses the file outright, failing to open 96.7\% of the same documents; its zero-yield rate of 96.8\% is almost exactly its open-failure rate.
One extractor logs these as successes with empty output, the other as parse errors.
The gap between the engines is concentrated in the 59.3\% of truncated documents that are not linearized; Appendix~\ref{app:lin} breaks recovery down by linearization.

Applying each engine's recovery to the exposure in Table~\ref{tab:trunc}, the loss $E\,(1-r_e)$ is $63.08\%\times(1-0.114)\approx56\%$ of the corpus's text under PyMuPDF and $63.08\%\times(1-0.014)\approx62\%$ under PDFium; we report the range 55--62\%.
The product multiplies an exposure counted in Tika tokens by a recovery counted in whitespace tokens.
Since every truncated document has both counts, we re-weight each by its Tika count instead.
The corpus figure moves by 0.42 points under PyMuPDF (to 55.3\%, the lower end of the range) and by zero under PDFium.
The two counts agree closely per document (median ratio 1.01--1.02).
They diverge only for writing systems without whitespace word separators, such as Japanese and Chinese, which is a property of the corpus rather than of truncation.
OCR is not an alternative route: a page can be rasterised only if its page object and content stream lie inside the fragment, in which case its text layer is extractable as well, and Poppler, the renderer behind the usual \texttt{pdf2image} OCR front end, refuses truncated files as PDFium does.
Re-fetching from origin, as CCpdf and FinePDFs did, is a remedy for the crawl rather than an extraction method, and it is what \S\ref{sec:cap5} argues for.
One caveat does remain: we count tokens rather than reading them, and a token count cannot distinguish text that a repair path recovered cleanly from reconstruction artefact, so ``recovers more'' must not be read as ``recovers better''.

\subsection{The \texorpdfstring{5\,MiB}{5 MiB} cap recovers little of what it still exposes}
\label{sec:cap5}

Raising the cap to 5\,MiB (Table~\ref{tab:trunc}) leaves 30.19\% of tokens \emph{exposed}; we tested whether they are \emph{recoverable}, since every document truncated at 5\,MiB is also truncated at 1\,MiB and only the cut point moves between the runs.

On the 316{,}174 documents PyMuPDF processed at both caps, recovery rises from 3.3\% to 13.2\% of tokens, and 21{,}203 documents (6.7\%) go from yielding no text at all to yielding some. PDFium moves from 0.1\% to 2.4\% on the same documents.
A five-fold larger prefix therefore multiplies recovered text roughly four-fold, but still leaves 77.6\% of these documents (97.0\% under PDFium) yielding no text.
The change halves the exposure; what it leaves behind remains almost entirely unreadable.

The benefit is also concentrated immediately above the cap (Table~\ref{tab:cap5}, Appendix~\ref{app:cap5}): 74\% of the rescued documents lie between 5 and 10\,MiB, while above 25\,MiB the larger cap is worth 2--3 percentage points.
This is what a fixed prefix must do, 5\,MiB is half of a 10\,MiB file and a twentieth of a 100\,MiB one, and it holds under both engines.
Raising the cap further has sharply diminishing returns per additional TiB of archive: the documents that dominate the remaining token mass are those a larger fixed prefix helps least.

Both effects reflect the cut rather than the run: token counts from the whole documents agree exactly for all but 0.036\% of paired documents under both engines (a shared, harness-imposed deadline).
Recovery is not monotonic, however: 800 documents (0.25\%) yield text at 1\,MiB and none at 5\,MiB, so repair is sensitive to \emph{where} a file stops, but the effect is 26$\times$ rarer than the reverse and does not disturb the aggregate.

\section{Conclusion}
Counted in documents and counted in text, the studied corpus \texttt{CC-MAIN-2021-31-PDF-UNTRUNCATED} is two different corpora: the units diverge by up to 11$\times$ per category, and Common Crawl's cap affected 23\% of documents but left 55--62\% of the text unrecoverable to text-layer parsers.
Corpus statistics should therefore be reported in both units; every rate in this literature (coverage, filter drop, OCR routing) is per-document or unreported.
Any filter that selects on size, such as PDFA's 100\,MB and 500\,ms cuts, FinePDFs' OCR router or the crawler's cap, removes far more text than documents, because size is where the text is; its effect should be reported in tokens.
For Common Crawl, \S\ref{sec:trunc}--\ref{sec:cap5} quantify the 1\,MiB\,$\rightarrow$\,5\,MiB change: it halves the exposure, but little of what it still truncates is recoverable; the remaining exposure calls for a size-aware fetch policy for large PDFs, not a higher cap.

\section*{Limitations}

Every figure describing corpus \emph{composition} and truncation \emph{exposure} is computed over all 7.9M documents from the shipped metadata.
The recovery figures cover the 1.2M truncated documents both extractors successfully processed, 67\% of the 1.83M Common Crawl truncated.
The shortfall is concentrated in ten of our sixty-four PyMuPDF shards, which produced nothing because of technical errors in the extractor (e.g. out of memory, hanging); since shards partition the corpus by archive index, the loss is arbitrary with respect to document content, and per-shard recovery rates vary by about a percentage point.
It is nonetheless not a random sample, and we report no confidence intervals.
The metadata is third-party and derived \citep{safedocs2021}, and \texttt{producer}/\texttt{creator} strings are self-reported with 8.08\% unclassified, so the \TeX{} share is a lower bound in both units.
The page-count bins of Table~\ref{tab:shape} and the toolchain families are heuristics, not genre labels: a long tax form falls in the $>$50-page bin, and a thesis typeset in Word falls outside the \TeX{} family.
\texttt{page\_size} is recorded for the first page only, making the landscape category approximate. 
We study a single crawl snapshot, from before the policy change, and the only one for which re-fetched originals exist; the 5\,MiB figure is therefore a counterfactual computed on 2021 file sizes rather than a measurement of a current crawl.
Re-measuring exposure on a post-2025 crawl needs only the WARC metadata and is the obvious next check; we have not done it.
Finally, both extractors are text-layer parsers and we run no OCR arm. 
A page can only be rendered if its objects lie inside the truncated document, in which case its text layer is extractable as well, so OCR cannot recover more from the crawl than a text-layer parser (\S\ref{sec:sim}).

\section*{Acknowledgement} We thank DFKI (Deutsches Forschungszentrum für Künstliche Intelligenz GmbH) for supporting this work with computation and storage.

\bibliography{custom}

\clearpage
\appendix

\section{Recovery by document size}
\label{app:cap5}

Table~\ref{tab:cap5} breaks the paired recovery measurement of \S\ref{sec:cap5} down by document size.

\begin{table}[th]
\centering\small
\begin{tabular}{lrrrr}
\toprule
& & \multicolumn{2}{c}{recovered} & \\
\cmidrule(lr){3-4}
size & docs & 1\,MiB & 5\,MiB & gain \\
\midrule
5--10\,MiB   & 179{,}810 & 4.88\% & 22.44\% & +17.6 \\
10--25\,MiB  &  99{,}831 & 2.62\% &  8.97\% &  +6.4 \\
25--50\,MiB  &  25{,}700 & 1.64\% &  4.47\% &  +2.8 \\
50--100\,MiB &   8{,}683 & 1.36\% &  3.28\% &  +1.9 \\
$>$100\,MiB  &   2{,}150 & 1.01\% &  2.96\% &  +2.0 \\
\bottomrule
\end{tabular}
\caption{What raising the cap buys, by document size (PyMuPDF, gain in percentage points). The same 316{,}174 documents are cut at both points; the benefit is concentrated immediately above the cap.}
\label{tab:cap5}
\end{table}

\section{Recovery by linearization}
\label{app:lin}

The gap is not spread evenly across the corpus.
A \emph{linearized} PDF (Adobe ``Fast Web View'') places its first page and a cross-reference table at the front of the file so that a partial copy is still renderable; 40.7\% of the documents here are linearized.
Under PyMuPDF those files recover 1.1\% of their tokens against 18.5\% for the non-linearized rest, while under PDFium it is 1.1\% against 1.5\%.
Note that the gap persists when files of similar size are compared, so it is not a size effect.
The difference between the two engines therefore lives in the non-linearized population, and the Fast Web View structure, designed to keep partial files usable, is the one from which least is recovered.

\end{document}